# A Target Detection Algorithm in Traffic Scenes Based on Deep Reinforcement Learning

Xinyu Ren*

School of Information Engineering, Chang'an University, 2021224006@chd.edu.cn

Ruixuan Wang

School of Transportation Engineering, Chang'an University, wangrx0818@126.com

This research presents a novel active detection model utilizing deep reinforcement learning to accurately detect traffic objects in real-world scenarios. The model employs a deep Q-network based on LSTM-CNN that identifies and aligns target zones with specific categories of traffic objects through implementing a top-down approach with efficient feature extraction of the environment. The model integrates historical and current actions and observations to make a comprehensive analysis. The design of the state space and reward function takes into account the impact of time steps to enable the model to complete the task in fewer steps. Tests conducted demonstrate the model's proficiency, exhibiting exceptional precision and performance in locating traffic signal lights and speed limit signs. The findings of this study highlight the efficacy and potential of the deep reinforcement learning-based active detection model in traffic-related applications, underscoring its robust detection abilities and promising performance.

## 1 INTRODUCTION

Target detection technology occupies a core position in the field of computer vision, and its main task is to identify and locate specific object entities from images or video streams. The technique involves not only identifying the categories of objects present in the image, but also precisely determining the spatial location of the objects, typically marked by drawing bounding boxes. These bounding boxes define the object's coordinates and dimensions on the image plane, providing a basis for further image analysis and interpretation.

Intelligent transportation systems [1] utilize target detection technology to optimize traffic management and planning. This reduces congestion and improves road usage efficiency by analyzing road usage in real time. In autonomous driving systems [2], accurate target detection is not only essential for identifying other vehicles, pedestrians, and road signs, but it is also crucial for vehicle navigation and safety decisions. Similarly, in traffic monitoring systems [3], target detection is used to track traffic flow and monitor traffic violations.

In the early days, target detection technology primarily relied on traditional methods. The core process of these methods can be summarized into three main parts: information area selection, feature extraction, and feature classification [4]. Traditional target detection often employs a sliding window strategy to scan the entire image and identify all possible target areas for information area selection [5]. However, this approach tends to generate a large number of redundant windows, significantly increasing the computational load and making it challenging to meet real-time performance requirements.

In traditional target detection, manually designed features are crucial. The Haar-like algorithm [6] is a fast feature extraction method based on integral images, effectively detecting image features such as edges, lines, and patches. However, this algorithm is not robust to changes in illumination and pose, often resulting in poor performance. The HOG algorithm [7] uses the histogram of gradients or edge directions to describe an object's shape characteristics. It exhibits a degree of illumination and scale invariance but may still struggle with recognition in complex scenes due to pose variations and occlusions, leading to degraded performance. The SIFT algorithm [8] identifies extreme points in scale space and extracts their descriptors, offering good scale, rotation, and brightness invariance but at the cost of high computational complexity, which can hinder efficiency in complex applications.

After feature extraction, a classifier is required to categorize and identify the features. For instance, the support vector machine (SVM) [9] classifier aims to find an optimal hyperplane in the feature space to maximize the margin between positive and negative samples, thus achieving effective classification. However, SVM's computational complexity can become prohibitive with large-scale datasets, affecting practical application outcomes. The Adaboost algorithm [10] builds a strong classifier by iteratively training and combining base classifiers with certain weights. Yet, Adaboost is highly sensitive to noisy data and outliers, potentially compromising the training process.

As the application of big data continues to deepen across various industries, traditional algorithms that rely on manual feature construction are no longer sufficient for the needs of the big data environment. Consequently, researchers have shifted their focus to deep neural networks, particularly convolutional neural networks (CNNs). Deep learning-based target detection algorithms can be primarily categorized into two types: two-stage algorithms that rely on region proposals and one-stage algorithms that rely on regression [11].

The R-CNN algorithm [12] is a leading example of the two-stage target detection approach. It was the first algorithm to successfully apply a CNN to the task of target detection. R-CNN works by extracting candidate regions from the input image and then using a CNN to classify each region, achieving target detection. However, R-CNN requires separate feature extraction for each candidate region, resulting in low computational efficiency. The Spatial Pyramid Pooling Network (SPP-Net) [13] addresses the issues of repetitive feature calculation and the fixed input image size found in R-CNN by incorporating a spatial pyramid pooling layer, thus improving the efficiency of target detection. Despite this, SPP-Net is unable to perform end-to-end training, which hampers further performance enhancements. Fast R-CNN [14] significantly improves both the performance and efficiency of target detection by simplifying the SPP layer, introducing the RoI Pooling layer, a joint training strategy, and a Softmax classifier. Nevertheless, its process for generating candidate regions still depends on traditional object extraction algorithms, which somewhat limits its detection speed. Building on this, Faster R-CNN [15] achieves the sharing of convolutional features through the introduction of a region proposal network and an anchor mechanism, greatly enhancing the model's speed and rendering it the first near-real-time target detection algorithm.

The one-stage target detection algorithms are more efficient than their two-stage counterparts and can meet real-time requirements, but they are generally not as accurate. The one-stage approach is exemplified by the YOLO (You Only Look Once) [16] algorithm. YOLO shifts the perspective of target detection from a classification to a regression problem, pioneering a new direction in the field. In recent years, numerous scholars have developed various iterations of the YOLO algorithm and have applied them across a broad range of fields. For instance, the enhanced YOLOv5 network [17] integrates the feature pyramid model AF-FPN to improve the detection performance of multi-scale targets.

In recent years, target detection algorithms based on deep reinforcement learning have been extensively studied within the field of computer vision and have demonstrated significant potential. Researchers have applied deep reinforcement learning to target detection [18], utilizing a dynamic attention mechanism to train the DQN network. This allows the network to progressively adjust and scale the focus area on the image until the target object is located. However, the target detection accuracy in this experiment remains low, and aspects such as the reward function, action space design, and state representation require further refinement.

Although object detection algorithms that employ deep reinforcement learning have shown promising results in certain tasks, these methods still encounter numerous challenges. These include dealing with unstable rewards and complex state spaces, as well as improving training stability, performance, and efficiency. As a result, the development of target detection algorithms based on deep reinforcement learning continues to be a vibrant area of research.

## 2 ANALYSIS OF THE TRAFFIC OBJECT DETECTION PROBLEM

The target detection algorithm designed in this experiment differs from common deep learning-based detection algorithms, such as YOLO, which typically divide the image into grids and traverse these grids. Instead, based on a supervised learning approach, this detection problem is treated as a regression task to predict class probabilities and object coordinates. The model proposed in this study employs a top-down search strategy for detection, initially analyzing the entire scene globally before progressively narrowing the search area to locate the specific position of the target object. This is achieved by performing a series of transformations on an observation area frame that initially covers the image globally. Through repeated iterations, the frame is refined to fit closely around the target. This method does not rely on a single structured prediction technique but rather employs a dynamic attention mechanism strategy that adjusts the observation frame in response to the current area's content, gradually honing in on the target's location.

Figure 1 illustrates the model's iterative process. The blue rectangular frame around the image's initial part indicates the default observation area demarcated by the model, encompassing the entire global image area. The blue frames in the subsequent parts of the image show the model's observation area during the iteration process. These frames, marked at a specific step and the final chosen step, increasingly focus on the area containing the target traffic light. As the iterations proceed, the model continuously shifts the observation frame towards the light's actual coordinates. During this iterative process, the frame's position in the next step is contingent upon its current location and the action to be taken at that moment. Once the frame progresses to the successive step, the process is repeated, enabling further refined decisions.

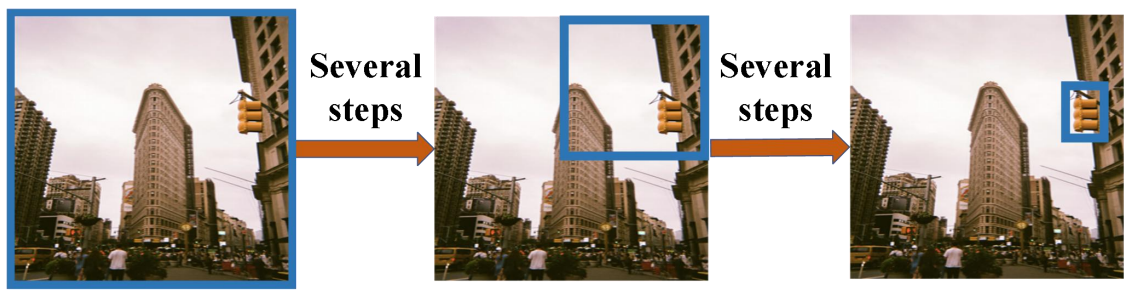

Figure 1: Iterative process of the model.

This process can be formalized as a Markov decision process (MDP) [19], wherein the state is represented by the position of the current observation area box along with several historical actions. The action is defined by the decision to change the position of the observation area box, and the reward is determined by how well the observation area box fits the target. Under this framework, the model's objective is to discover a strategy that maximizes the cumulative reward from the initial state to the goal state through a series of actions. Consequently, this study introduces a reinforcement learning algorithm, which enables the model to incrementally optimize its decision-making through continuous interaction with the environment. At each step, the reinforcement learning algorithm updates its strategy based on the current state and rewards, thus progressively refining the position and size of the observation area box. This iterative improvement allows the observation area box to locate the target object more accurately. Given that the state includes image data, the state space is relatively complex, comprising a high-dimensional vector of continuous values. Therefore, the experiment employs a deep Q-network (DQN) as the primary algorithm, leveraging the concept of deep reinforcement learning to fulfill the task of traffic object detection.

## 3 DESIGN OF THE TARGET DETECTION MODULE BASED ON DEEP REINFORCEMENT LEARNING

The state in the target detection's Markov decision process is defined by the image area within the current observation area frame and some historical actions. An action is an operation on the observation area frame selected from the available action space. The reward function evaluates the impact on the test results after executing an action. For each step, the model chooses an action given the state and executes it. Depending on the action performed, the model receives a reward and transitions to a new state. This reward is used to assess the effects of the selected actions and provide feedback for subsequent decisions. The goal of the model is to identify a policy that maximizes the expected cumulative reward.

The modules and their respective functions included in building a target detection model are as follows:

**Dynamic Decision-Making Process Module.** This module first preprocesses the collected image datasets, annotates the target locations of traffic objects within the images, saves the data required for deep reinforcement learning tasks, and constructs an environment for deep reinforcement learning. Secondly, it designs an appropriate action space, state space, and reward function for the target detection task to lay the groundwork for subsequent model training and prediction steps.

**DQN Network Building Module.** This module first establishes a suitable network structure based on the characteristics of states, actions, and other elements in the deep reinforcement learning task so that the DQN network can better interpret the environment's state. Secondly, the training process for the DQN network is outlined, including the specific steps of model training on the training set, the updating of DQN network parameters, and parameter selection. Finally, a process is designed for the network to detect traffic objects after the image data is input into the trained DQN network, enabling the model to perform its functions and prepare for subsequent model testing.

**DQN Network Test Module.** This module selects appropriate evaluation metrics to assess the model's performance and determine whether it can accurately label traffic objects in images from the test set.

## 4 DESIGN OF THE DYNAMIC DECISION-MAKING PROCESS

### 4.1 Environment

For traffic object detection, the model's image inputs are treated as an 'environment', as determined by a requirements analysis for addressing the detection task through reinforcement learning methods. We assembled a dataset of 5,000 traffic images, captured under diverse environmental conditions, from various perspectives, and at different proximities. Each of them contains a traffic light or a speed limit sign target. The annotation process entailed marking each image with relevant features, such as image names, pixel RGB values, and target coordinates. This manual annotation delineated the required traffic objects in each image, resulting in the storage of image paths, names, sizes, and details of the observed areas locally.

Following the acquisition of image data through designated paths and names, we applied a normalization process to the pixel RGB values, scaling them to a range of [0,1]. This scaling optimizes the feature extraction process for neural networks. The image data for each traffic object category was then organized into a set of tensor objects. These tensors encapsulate the normalized pixel data, label coordinates, and image dimensions, with the image name serving as the index key. To ensure standardized data input for model training, all tensor objects were reshaped to a consistent dimension of 224x224 pixels.

The traffic image dataset was divided into train and test sets at a 9:1 ratio.

### 4.2 Actions

We utilize a discrete action space, where the agent's task involves adjusting the observation area frame through translational or scaling movements, measured in pixel units. The action space designed for this experiment comprises nine distinct actions. The action space is defined as A = {0,1,2,3,4,5,6,7,8}, where the action '0' indicates maintaining the current position of the observation frame. As is listed in Table 1, actions '1' through '4' represent translational shifts of the frame, and actions '5' through '8' correspond to scaling adjustments. These actions alter the frame's position or dimensions by a pixel distance that is proportional to the image's dimensions, a ratio known as the 'scaling factor'. We set the scaling factor at 0.03. This factor determines the extent of each translational or scaling change in relation to the overall size of the image. After any action, coordinate adjustments are made to keep the frame within the image's valid range: any resulting negative coordinate values are reset to 0, and any values exceeding the maximum of 224 are capped at 224.

Table 1: Evaluation results on the test set

| ID | Action | Description |
|---|---|---|
| 0 | Select | Select the position of the observation area box |
| 1 | Move Right | Increase both the maximum and minimum x-coordinates by 224*3% |

| ID | Action | Description |
| --- | --- | --- |
| 2 | Move Left | Decrease both the maximum and minimum x-coordinates by 224*3% |
| 3 | Move Up | Decrease both the maximum and minimum y-coordinates by 224*3% |
| 4 | Move Down | Increase both the maximum and minimum y-coordinates by 224*3% |
| 5 | Enlarge | Decrease both the minimum x-coordinate and y-coordinate by 224*3%, and increase both the maximum x-coordinate and y-coordinate by 224*3% |
| 6 | Shrink | Increase both the minimum x-coordinate and y-coordinate by 224*3%, and decrease both the maximum x-coordinate and y-coordinate by 224*3% |
| 7 | Widen | Increase the minimum y-coordinate by 224*3% and decrease the maximum y-coordinate by 224*3% |
| 8 | Narrow | Increase the minimum x-coordinate by 2243% and decrease the maximum x-coordinate by 2243% |

### 4.3 State

We adopt a bifurcated approach to constructing the state space, integrating both continuous and discrete elements. The state of the agent at any given timestep is articulated as $s = (f, h)$, where $f$ represents the feature vector extracted from the observed area frame, and $h$ embodies the vector detailing historical actions, conceived in accordance with sliding window theory.

Leveraging the concept of transfer learning, the experiment employs the convolutional layers and the penultimate fully connected layer of the VGG16 network pre-trained on the ImageNet dataset, as the feature extractor. In this experiment, the network does not undergo further training with image data from the observation area. As a result, the Dropout layers within the network are deactivated, and the batch normalization layers operate using running averages and variances to ensure accurate model outputs. This approach results in a continuous state space, with the dimensionality of the resulting state vector fixed at 4096.

At the beginning of each episode, a two-dimensional vector with dimensions (9,9) is created, uniformly initialized to one. Following an action, the executed action is converted into a One-Hot encoded vector of length 9, characterized by zeroes except for the index corresponding to the performed action, which is assigned a value of one. This vector is then enqueued at the beginning of the sliding window queue, while the oldest vector is removed. Subsequently, the vector is flattened to yield a dimensionality of 81.

The final step involves combining the two feature vectors of continuous and discrete horizontally to form a comprehensive state vector with a dimensionality of 4177. This vector serves as the agent's state representation during each timestep of the reinforcement learning process.

### 4.4 Reward

A reward function is instrumental in guiding the agent's actions toward achieving an objective or optimizing a certain criterion within the environmental context. Owing to its reliance on comparative analysis with pre-established benchmarks, the reward function is typically approximated during the training phase.

We define the reward function in relation to two principal agent actions: the selection of an observation area frame and its subsequent adjustment, either through movement or scaling. Following each action, the computation of the reward function splits depending on the nature of the action, including selection or adjustment of the frame.

### 4.4.1 Selection Action Reward.

Upon the selection action, the agent's reward depends on the Intersection-over-Union (IoU) metric between the pre-labeled target's actual position and the predicted position of the observation frame. A higher IoU ratio indicates a better decision, warranting a correspondingly higher reward. Let $o$ be the observation area frame and $g$ be the real area frame of the target object. The IoU between $o$ and $g$ is defined as (1):

$$IoU(o,g) = area(o \cap g) / area(o \cup gr) \qquad (1)$$

The reward function is built around a predefined IoU threshold $\tau$. If the IoU exceeds this threshold, the agent receives substantial positive reinforcement; conversely, if it is below, the agent experiences significant negative reinforcement. The reward for selecting is given by (2):

$$R_a(s_t, s_{t+1}) = \begin{cases} \max(3, (IoU(ob, gr) - \tau) * v) + t/500, & IoU(ob, gr) \geq \tau \\ \min(-3, (IoU(ob, gr) - \tau) * v) + t/500, & IoU(ob, gr) < \tau \end{cases} \qquad (2)$$

In (2), $a$ represents the action selected in this step, and $s_t$ represents the status of step $t$. $\tau$ is the intersection-to-union ratio threshold, and $v$ is a coefficient that amplifies the difference between the actual intersection-to-union ratio and the threshold. In this experiment, $\tau$ is set as 0.66.

We set $\tau$ to 0.66, indicating that a positive reward is granted only if the IoU exceeds 0.66 following a "select" action. The parameter $v$ is established at 20. The formula accounts for the influence of the time step $t$ on the reward. This methodology aligns with a prevalent reinforcement learning principle, which favors choosing shorter step sequences to achieve the goal, thereby minimizing instances where the observation area box repeats the same sequence of actions without successful selection. Additionally, after executing a "select" action, the reward includes an extra incentive proportional to the number of steps $t$. This means that the intelligent agent is awarded a higher reward for completing the task in fewer steps.

### 4.4.2 Non-Selection Action Reward.

For actions other than "select", the reward is proportional to the improvement in the IoU after an action that modifies the observation frame's position. The reward for transitioning from state $s_t$ to $s_{t+1}$ is given by (3):

$$R_a(s_t, s_{t+1}) = sign(IoU(ob_{t+1}, gr) - IoU(ob_t, gr)) - t/500 \qquad (3)$$

The given formula indicates that if the Intersection over Union (IoU) increases from state $s_t$ to $s_{t+1}$, the first term of the reward function is assigned a value of 1. If not, this term is assigned a value of -1. This design precisely reflects the quality of the action selected in that step, meaning the agent is penalized when its actions cause the observation area box to move away from the target; conversely, the agent receives a positive reward when the action brings the box closer to the target. Additionally, the formula accounts for the impact of the step count $t$ on the reward. After the agent performs any non-"select" action, a penalty term negatively proportional to the number of steps $t$ is introduced into the reward. This implies that the more steps the agent takes without completing the task, the lower the reward it will receive. Similar to the second term in the "select" action formula, the objective of this term is to minimize the instances where the observation area box repeats the same action sequence without successfully selecting the area.

## 5 DQN NETWORK CONSTRUCTION

### 5.1 Network Structure Design

In this experiment, the agent updates its state starting from step $t = 0$ in each training round. The state at each step is defined by the image features and 9 items extracted by the feature extractor, which is based on the VGG16 architecture, within the observation area frame at that step. Historical actions are recorded using One-Hot encoding and are stored in a sliding window with a length of 9. Consequently, the Long Short-Term Memory-Fully Convolutional Network (LSTM-FCN) [20] is employed as the primary architecture for both the policy network and the target network. The detailed structure of the LSTM-FCN model is depicted in Figure 2.

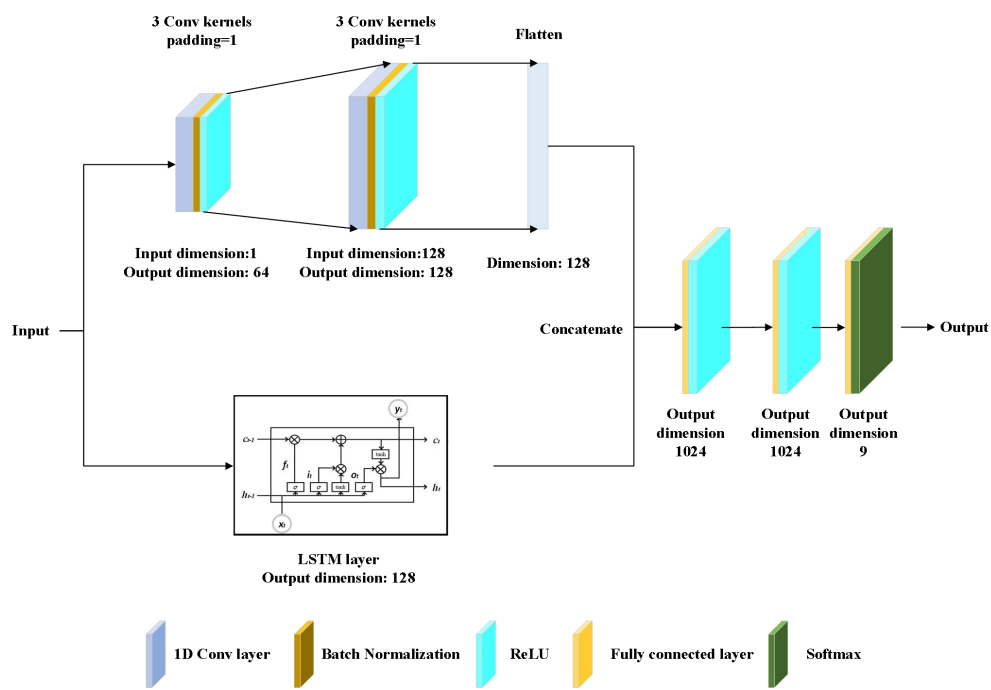

Figure 2: Structure of the DQN network.

The LSTM [21] used in the model is a type of recurrent neural network (RNN) designed to process sequential data. In a round of this deep reinforcement learning experiment, the state consists of a sequence of time steps. LSTM networks are well-suited for processing data with such sequential characteristics, as they can retain past information and utilize it as needed. A key advantage of LSTM networks is their ability to learn long-term dependencies, which is crucial for reinforcement learning tasks where the agent's current actions may influence its future states and rewards. By employing LSTM networks, agents can better understand these long-term dynamics.

The Fully Convolutional Network (FCN) [22] can handle high-dimensional inputs and transform them into low-dimensional feature vectors, simplifying the processing by the fully connected layer. This layer acts as a classifier, accurately mapping the combined output of the LSTM and FCN to the probabilities of selecting one of 9 actions. This enables the DQN to precisely select the most appropriate action.

### 5.2  Network Update Process

The refinement of policy and target networks predicated on the CNN model is crucial following each action of the agent and the corresponding reward computation. The policy network determines the selection of actions, whereas the target network offers a stable baseline for Q-values to ameliorate training stability. This update mechanism encompasses the following procedural steps.

**Sampling from Experience Replay.** The initiation of the process involves the extraction of a random assortment of 100 experiences from the experience replay buffer. This buffer is a repository that holds the agent's accumulated experiences, which include states, actions, rewards, and subsequent states. The stochastic nature of this sampling is key to maintaining the assumption that the samples are independent and identically distributed, which in turn breaks temporal correlations and improves the efficiency of data use.

**Batch Construction.** A batch is formed from these randomly selected samples, containing state, action, and reward information. This collective data is then employed in the execution of gradient descent and in advancing the updates of network parameters.

**Q-value Calculation via Policy Network.** The policy network receives the state and action data as inputs to calculate the Q-values for each pairing of state and action. In parallel, the target network is used to determine the maximum Q-value for every non-terminal state. This operation helps to differentiate the functions of the two networks in the processes of action selection and generation of stable targets.

**Target Q-value Estimation.** The target Q-value is derived using the Bellman equation, which amalgamates the immediate reward with the discounted Q-value of the following state. The discount factor, set at 0.9, adjusts the importance placed on future rewards during this calculation.

**Loss Function Computation.** The loss function is based on the mean squared error, calculated by comparing the predicted Q-values with the target Q-values. This evaluation of the loss function follows the principles of cross-entropy.

**Backpropagation and Parameter Update.** Backpropagation is implemented, and parameter updates are conducted using the Adam optimizer. This optimizer starts by resetting gradients, then proceeds to calculate the gradient of the loss function. Following this, the Deep Q Network's (DQN) parameters are updated in line with the calculated gradients.

### 5.3  Network training

The primary goal of model training is to enable an agent to effectively detect and accurately annotate objects within images. This protocol incrementally improves the agent's capability through carefully designed stages.

**Initial Training Configuration.** Parameters for the training regimen are established, stipulating the total number of iterations for each image in the dataset and the maximum allowable steps in each iteration. The image, along with its associated ground truth for object locations, is extracted to act as the training example. This sets the observational frame to cover the entire image initially.

**Iterative Action-Selection Cycle.** An iterative cycle commences with a step counter at zero. A series of actions are performed on each image until either a terminal action ("select") is taken or the step threshold is met. The agent's response varies depending on the action's nature. When a "select" action occurs, the action is recorded, the current coordinates of the observational frame are computed, and the maximum IoU score is calculated relative to both the recorded actions and the ground truth. The corresponding reward is then derived, indicating the end of the current iteration. For non-"select" actions, the historical action registry is updated, new observational frame coordinates are deduced, the image is cropped accordingly, and a new state is generated based on the image's feature vector and the history of actions within a sliding window. The optimal action to move or resize the frame is identified, the IoU score is calculated as before, and the resulting reward is computed.

**Experience Replay and Network Optimization.** Throughout this phase, each action taken by the model is followed by logging the current state, action, subsequent state, and reward into the experience replay buffer. The model optimization function is then invoked to update the policy network. After completing training for each image, the parameters of the policy network are transferred to the target network, and the current model configuration is archived.

The training incorporates an $\epsilon$-greedy strategy to balance exploration with exploitation. With this strategy, the agent selects actions randomly with a probability of $\epsilon$, and with a complementary probability of $1 - \epsilon$, it chooses the optimal action as estimated by the policy network.

At the beginning of training, $\epsilon$ is set to 1, which encourages purely random action selection. For the initial five iterations, $\epsilon$ is reduced by 0.18 after each iteration, promoting significant exploration early on. As training progresses, the tendency for exploration is gradually reduced, decreasing the likelihood of random action selection to 10% after five iterations and increasing the reliance on the knowledge acquired previously.

The training is limited to 15 rounds, each with no more than 100 steps.

## 6 PREDICTION

After saving the trained DQN network locally, the model can load the network to detect target locations in images beyond the training set. During the initialization of the training session, the agent is not informed about the number of targets in the image and does not know whether there are other targets outside the selected observation area. Therefore, we design a separate solution for images with multiple targets.

**Target Detection in a Single Image.** In this method, after reading an image, the model follows a process similar to the single-target detection method previously described to detect a target in the image. The DQN network first observes the image and selects the most prominent target. This step is achieved through a preset state-action strategy.

**Masking the Selected Target.** Instead of ending the program and outputting the result immediately after a target is selected, the 'select' action is followed by masking the area within the observation box with the average pixel value of the image, effectively concealing the first target.

**Reset and Repeat Detection.** Subsequently, the step count $t$ is reset to 0, and the single-target detection process is repeated. This process continues in a loop until the model can no longer detect new targets in the image.

**Output of Final Detection Results.** After all targets have been successfully detected, the program outputs an image with multiple blue rectangular boxes as the final detection result. These blue rectangles mark all the targets detected in the image. This result is saved in a folder.

The model's process of detecting the positions of traffic object targets in the image is illustrated in Figure 3, with the blue rectangles representing the target area boxes.

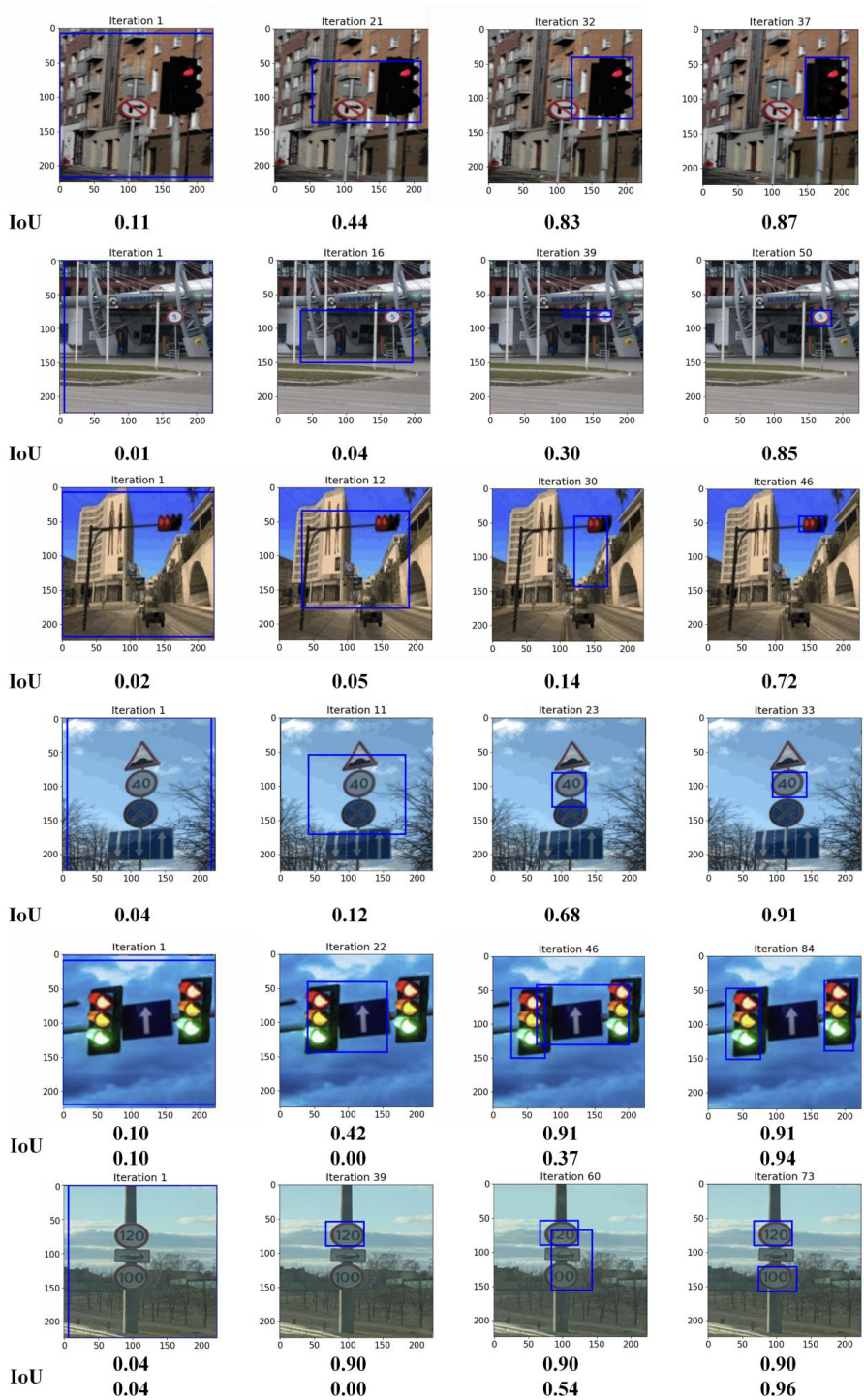

Figure 3: Process of target detection.

## 7 EXPERIMENTAL RESULTS

After inputting the data from the test set into the well-trained and saved DQN network, the model predicts the positions of specified traffic objects in each image of the test set following the above prediction process and marks them with blue rectangular frames. For each image, after the prediction is completed, the predicted results are compared with the true values of the target positions previously annotated in the training set, and the IoU between the predicted results and the actual target positions is calculated. In this experiment, a prediction is defined as correct if the IoU is greater than or equal to 50%, meaning the prediction can determine the position of the specified traffic objects in the image with reasonable accuracy.

Based on the above analysis, it is evident that this problem is a binary classification issue. In this experiment, two evaluation metrics are introduced: Average Precision (AP) and Recall, to assess the performance of the model in delineating traffic lights and speed limit signs respectively. Each image from the test set is input into the DQN network to calculate the IoU between the predicted and the actual bounding boxes. If the IoU exceeds the 50% threshold, this indicates that there is sufficient overlap between the predicted and actual bounding boxes, and the prediction is considered correct, counting as a True Positive (TP). Otherwise, the prediction is wrong, accumulating as a False Positive (FP).

In this experiment, since the test set contains no images without objects, the sum of True Positives (TP) and False Negatives (FN) equals the total number of samples. To avoid redundancy, metrics such as accuracy and precision are not recalculated. The evaluation results of the model on the test set are shown in Table 2.

Table 2: Evaluation results on the test set

| ID | Traffic Object | Average Precision | Recall |
|---|---|---|---|
| 1 | Traffic lights | 0.789 | 0.875 |
| 2 | Speed limit signs | 0.949 | 0.966 |

To verify the model's performance in detecting traffic object targets, the number of steps taken by the model to successfully detect traffic lights and traffic speed limit signs 200 times each in the images was recorded. The distribution of the number of steps for detecting traffic light and speed limit sign targets is shown respectively in Figure 4, while statistical indicators of performance in detecting traffic lights and traffic speed limit sign image targets are shown in Table 3.

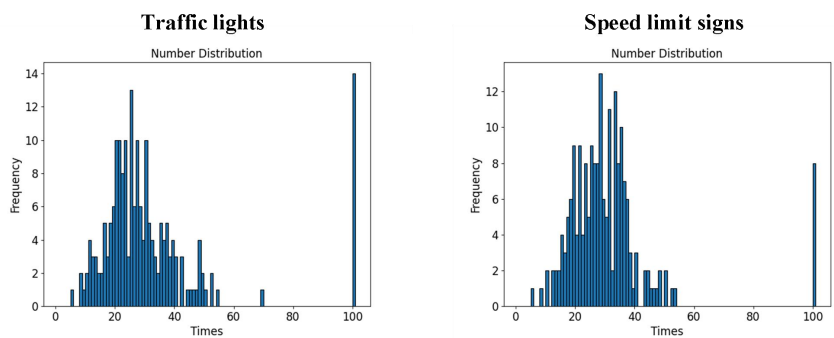

Figure 4: Number of steps for detecting traffic objects.

Table 3: Statistical indicators of performance

| ID | Indicator | Traffic lights | Speed limit signs |
|---|---|---|---|
| 1 | Times of successful detection | 200 | 200 |
| 2 | Times of triggering | 186 | 192 |
| 3 | Percentage of triggering | 93.00% | 96.00% |
| 4 | Times of reach $t$=100 | 14 | 8 |
| 5 | Percentage of reach $t$=100 | 7.00% | 4.00% |
| 6 | Mean value of $t$ when completing detection | 32.02 | 30.82 |
| 7 | Medium value of $t$ when completing detection | 27 | 28 |
| 8 | Minimum value of $t$ when completing detection | 5 | 5 |
| 9 | Maximum value of $t$ when triggering | 69 | 53 |

It is demonstrated that the model identifies the location of targets in the majority of images by executing the "select" action. For the minority of images where the model does not choose the "select" action within a limited number of steps, there is still a certain probability that the observation target box will achieve an IoU of 50% with the true coordinates of the target. The model completes the detection of traffic object targets on most images within 20 to 40 steps. Upon examination, it is observed that this is because, in the collected image dataset, the area occupied by most traffic lights or traffic speed limit signs constitutes about 2 to 15% of the entire image area. Meanwhile, each non-"select" action alters the size or position of the observation area box by about 3% to 5.91% (1 - 0.97*0.97/1) of the image area.

## 8 CONCLUSION AND FUTURE WORK

In this article, we introduce an active detection model based on deep reinforcement learning for detecting traffic objects in real-world scenarios. First, we collect and label a set of traffic image datasets containing traffic objects. Then, using the deep reinforcement learning approach, we construct the model and define the action space, state space, and reward function in detail. Subsequently, a deep Q-network (DQN) based on the LSTM-FCN model is established. After processing the image and the historical action-related feature data, the model selects a movement so that the observation area box moves closer to the traffic object's location, thereby achieving the detection target. The experimental results verify the model's effectiveness, indicating that it has accurate target detection capabilities and good performance.

In the research domains of small object detection and multi-object detection, our method has exhibited certain limitations. The study suggests that if the intelligent agent's action design involves large amplitude adjustments, it may lead to the target bounding box including too much background information. Conversely, if the adjustment amplitude is too small, it requires additional exploration steps, which not only slows down the localization process but may also cause instability in the actions during small object detection, such as swaying or backtracking. Moreover, when faced with images containing multiple similar or identical targets, the intelligent agent may find it challenging to effectively differentiate and extract the visual features of each target, which could result in the exploration task being interrupted due to exceeding the maximum number of steps. These challenges highlight the performance limitations of the method when dealing with specific detection problems.

However, this innovative algorithm employs a dynamic attention mechanism that actively searches for targets from a top-down perspective, focusing computational effort on areas of interest, in contrast to traditional deep learning-based object detection algorithms such as YOLO, which process the entire image uniformly across multiple regions. This targeted approach of the proposed method may lead to more efficient detection, particularly in complex scenarios, suggesting a promising avenue for future research and application.

Due to the constraints of the researchers' time and capabilities, the exploration of these optimization spaces is deferred to future research work.